\documentclass{article}

\usepackage{microtype}
\usepackage{graphicx}
\usepackage{subfigure}
\usepackage{booktabs} 
\usepackage[table]{xcolor}
\usepackage{hyperref}

\usepackage[accepted]{icml2025}

\usepackage{amsmath}
\usepackage{amssymb}
\usepackage{mathtools}
\usepackage{amsthm}

\usepackage[capitalize,noabbrev]{cleveref}

\usepackage{newfloat}
\usepackage{listings}
\usepackage{mathrsfs}
\usepackage{xcolor}
\usepackage{nicefrac}
\usepackage[noend]{algpseudocode}
\usepackage{multirow}
\usepackage{newtxtext}
\usepackage{algorithm}
\usepackage{longtable}

\theoremstyle{plain}

\theoremstyle{definition}

\theoremstyle{remark}

\DeclareMathOperator*{\argmin}{arg\,min}

\usepackage[textsize=tiny]{todonotes}

\icmltitlerunning{}

\begin{document}

\twocolumn[
\icmltitle{TeamFormer: Shallow Parallel Transformers with Progressive Approximation}



\icmlsetsymbol{equal}{*}

\begin{icmlauthorlist}
\icmlauthor{Wei Wang}{yyy}
\icmlauthor{Xiao-Yong Wei}{yyy}
\icmlauthor{Qing Li}{yyy}
\end{icmlauthorlist}

\icmlaffiliation{yyy}{Dept. of Computing, Hong Kong Polytechnic University, Hong Kong}

\icmlcorrespondingauthor{Xiao-Yong Wei}{
x1wei@polyu.edu.hk}

\icmlkeywords{Machine Learning, ICML}

\vskip 0.3in
]



\printAffiliationsAndNotice{}  

\begin{abstract}
The widespread ``deeper is better'' philosophy has driven the creation of architectures like ResNet and Transformer, which achieve high performance by stacking numerous layers. 
However, increasing model depth comes with challenges such as longer training times, higher inference latency, and impracticality on resource-constrained devices.
To address these issues, we propose TeamFormer, a shallow Transformer architecture designed for true parallelism in both structure and computation. 
By formulating standard Transformers as function approximators in closed-form, our theoretical analysis shows that their performance relies on inter-layer collaboration for progressive approximation, rather than depth itself. 
While deep Transformers enforce this collaboration through sequential designs, we demonstrate that such collaboration is not inherently tied to sequential structures.
TeamFormer removes the sequential constraint by organizing layers into parallel branches, enforcing inter-layer collaboration algorithmically. 
Specifically, we implement progressive approximation, ensuring that each new branch further reduces the loss from preceding branches, enabling faster convergence.
Extensive experiments validate TeamFormer’s effectiveness, outperforming standard Transformers like ViT. 
Moreover, TeamFormer supports up to $15.07\times$ model compression and facilitates model expansion for adaptive continuous learning.
Experimental results on multi-GPU deployment demonstrate that TeamFormer is $3.30\times\pm 0.86$ faster than widely used parallelism solutions such as FairScale.
These advancements stem from our closed-form formulation of Transformers based on the Universal Approximation Theorem, which not only explains the ``depth belief'' but also opens new avenues for designing efficient Transformer architectures. \texttt{https://(open-upon-acceptance)}
\end{abstract}


\section{Introduction}
\begin{figure}[htp]
\centering
\includegraphics[width=0.5\textwidth]{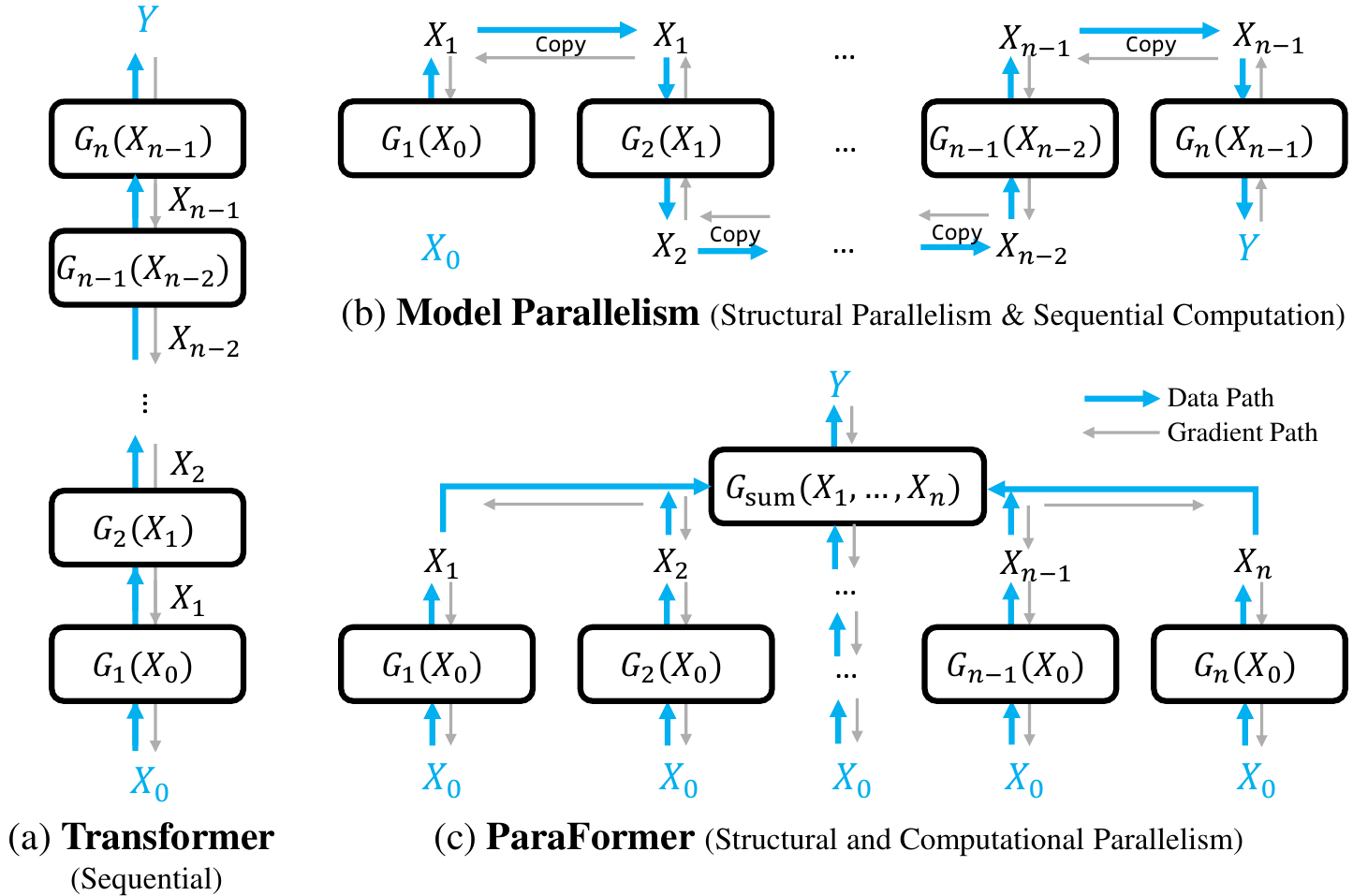}
\caption{Architectures of Transformer and its parallelism variants: (a) Original Transformer; (b) Model Parallelism; and (c) The Proposed \textbf{TeamFormer}.}
\label{fig:struc_comp}
\end{figure}
In recent years, the advancement of deep neural networks has largely been guided by the ``deeper is better'' philosophy \cite{Cohen2015OnTE,Delalleau2011ShallowVD}. 
As a result, leading architectures such as ResNet and Transformer typically achieve high performance by stacking numerous layers.
However, increasing model depth introduces significant challenges. 
Training times grow super-linearly, and inference latency becomes substantial, making deep models impractical for many applications.
For instance, clients who lack access to high-performance GPUs cannot feasibly deploy or run inference with large models released by major AI companies. 
Additionally, in situations requiring rapid response, such as autonomous driving or other real-time systems, this latency is often unacceptable.

To tackle these challenges, researchers have explored several optimization strategies, such as model parallelism \cite{Narayanan2021EfficientLL, Shazeer2018MeshTensorFlowDL}, data parallelism \cite{Dean2012LargeSD}, and dynamic networks \cite{Fedus2021SwitchTS}.
Consider model parallelism, which is among the most straightforward approaches. 
It reduces memory demands on individual devices by distributing different layers or parameter groups of a large model across multiple devices.
While promising results have been observed (see Figure~\ref{fig:struc_comp}b), this technique primarily achieves structural parallelism, not true computational parallelism. 
Data and gradients still flow sequentially through all layers as in the original architecture, and substantial overhead is introduced due to frequent copying operations between devices or branches.

In this paper, we study the mechanism within Transformers that drives their depth-oriented performance, and based on this understanding, we propose a shallow Transformer architecture called \textbf{TeamFormer} (Figure~\ref{fig:struc_comp}c), which achieves true parallelism in both structural and computational aspects.  
To achieve this, we have formulated the computation process of standard Transformers. 
Our theoretical analysis reveals that Transformers are progressive approximators to reduce the difference between the input and output (e.g., $\|\mathbf{X}_0 - \mathbf{Y}\|$, as shown in Figure~\ref{fig:struc_comp}a).  
This progressive process highlights that the existence of an approximation function $G_i()$ of the layer $i$ depends on the existence of all preceding functions (i.e., $G_1()$ to $G_{i-1}()$ in the lower layers). 
%
%
This analysis reveals that the advantage of depth lies not in the sequential structure itself, but in the enforced inter-layer collaboration for progressive approximation.  
Based on this insight, TeamFormer (Figure~\ref{fig:struc_comp}c) removes the sequential constraint and arranges the layers into parallelized branches. 
We enforce inter-layer collaboration for progressive approximation at the algorithmic level rather than through structural design, by incrementally adding new branches during training and ensuring that each new branch is introduced only after all preceding branches have been sufficiently trained (Figure~\ref{fig:prog_training}).

Our contributions are as follows:
\begin{itemize}
    \item \textbf{Closed-Form Formulation:}
    We reformulate the standard Transformer architecture by expressing the self-attention component of each block as the multiplication of a transformed attention matrix with a flattened vector of token embeddings through a series of matrix operations.
    When the result of this multiplication is fed into the feed-forward neural network (FNN) component, we found that the computation at each layer satisfies the conditions of the Universal Approximation Theorem (UAT) \cite{Cybenko1989ApproximationBS}, ensuring that each layer functions as a function approximator.
    This inter-layer collaboration establishes a foundation for the overall architecture’s optimization objective in closed-form.
    This formulation not only provides a theoretical basis for guiding parallelism but also offers a general framework for analyzing Transformers more comprehensively.

    \item \textbf{True Parallelism:}
    As illustrated in Figure~\ref{fig:struc_comp}c, the proposed TeamFormer achieves true parallelism, where data and gradients pass through corresponding branches independently. 
    All branches share the same input, $\mathbf{X}_0$, eliminating the need for heavy cross-branch copying. 
    This parallelism is both structural and computational, ensuring efficiency and scalability.

    \item \textbf{Flexible Model Compression and Expansion:}
Thanks to its parallel structure and progressive training approach, TeamFormer offers the flexibility to remove the final branch while allowing the remaining branches to function with only a limited drop in performance. 
This adaptability is beneficial in two key scenarios: 1) If clients lack sufficient computational resources to deploy a large-scale TeamFormer, they can remove later branches until the reduced model fits their constraints.
2) When data volume grows and an existing TeamFormer needs to be scaled up, new branches can be added and fine-tuned, without rebuilding the model from scratch.

    \item \textbf{Rationale Behind Depth Belief:} We identify that the performance gains achieved by increasing depth in Transformers are primarily due to the enforcement of inter-layer collaboration for progressive approximation, rather than depth itself. 
    This insight opens new opportunities for designing more effective and efficient Transformer architectures in the future. 
\end{itemize}

\section{Related Works}
\noindent\textbf{Efficiency Challenges and Optimization Methods in Deep Models:} Modern deep learning models, represented by architectures such as Transformer \cite{Vaswani2017AttentionIA} and ResNet \cite{He2015DeepRL}, have evolved into a wide range of variants \cite{Xie2016AggregatedRT, Tan2019EfficientNetRM, Devlin2019BERTPO, achiam2023gpt, touvron2023llama}. These models typically rely on stacking multiple layers to achieve high performance. However, this reliance on depth introduces significant challenges, including high computational cost, increased inference latency, and difficulties in deployment, especially on resource-constrained platforms.

Many edge devices, such as smartphones and IoT devices, struggle to handle the computational demands of large models, highlighting the need for lightweight and efficient solutions for such scenarios. To address these, various model acceleration techniques have been proposed. In the area of model compression, methods such as pruning \cite{Ma2023LLMPrunerOT, Xia2023ShearedLA}, quantization \cite{Jacob2017QuantizationAT, Lu2022AHA, Guo2021IntegerOnlyNN, Lin2023AWQAW}, and knowledge distillation \cite{Gou2020KnowledgeDA, Gu2021OpenvocabularyOD} aim to reduce model size and computational complexity. Additionally, dynamic networks, such as those based on conditional computation (e.g., Switch Transformer \cite{Fedus2021SwitchTS}), selectively activate parts of the model depending on input characteristics.

However, most of these approaches come at the cost of performance degradation or require retraining, and they primarily focus on reducing parameter count or skipping certain computations dynamically. Importantly, they do not fundamentally alter the sequential structure of the model. In contrast, this paper proposes a Structural and Computational Parallelism approach that redefines the computational flow from the ground up.

\noindent\textbf{Model Parallelization Techniques:} Current model parallelization strategies can be categorized into four types:
\begin{itemize}
\item  Data Parallelism \cite{Dean2012LargeSD}: Involves splitting the input data into mini-batches and distributing them across multiple devices (e.g., GPUs). Each device computes gradients on its own mini-batch, and the gradients are aggregated and synchronized on a primary device before updating the model parameters.

\item  Model Parallelism \cite{Narayanan2021EfficientLL, Shazeer2018MeshTensorFlowDL}: Splits the model across devices, particularly useful when a model is too large to fit on a single device. It is done manually by assigning different layers to different devices. During forward propagation, computation proceeds in the order of the assigned devices, while backward propagation follows the reverse order.

\item  Pipeline Parallelism \cite{Huang2018GPipeET, Park2020HetPipeEL, Li2021ChimeraET}: Also known as inter-layer model parallelism, it partitions the model into stages, allowing different stages to process data in a pipeline.

\item  Tensor Parallelism \cite{Narayanan2021EfficientLL}: Involves splitting matrix operations across devices, often requiring frequent inter-device communication.
\end{itemize}
Despite their effectiveness, these existing parallelization techniques are inherently pseudo-parallel, as computations still depend on a sequential data flow. None of them fully decouple computation from model structure. In contrast, TeamFormer achieves true parallelism through a branched architecture and progressive training, enabling parallel computation without data dependency.

\noindent\textbf{Transformer Architectures:} The Transformer architecture was first introduced by Vaswani et al. \cite{Vaswani2017AttentionIA} for machine translation tasks. Over time, its powerful representational capability has been widely recognized, especially in the context of large language models (LLMs). This success has led to the development of vision-based architectures such as ViT \cite{Dosovitskiy2020AnII}, which adapts the Transformer framework for image processing.

Transformers are prevalent across both modalities. In NLP, notable examples include BERT \cite{Devlin2019BERTPO} and LLaMA \cite{touvron2023llama}. In computer vision, models like DeiT \cite{Touvron2020TrainingDI} and Swin Transformer \cite{Liu2021SwinTH} have demonstrated strong performance.

The widespread adoption of Transformer highlights its effectiveness and flexibility. In this paper, we further analyze the theoretical advantages of Transformer and establish its mathematical foundation for parallelization, paving the way for truly parallel architectures like TeamFormer.

\section{Transformers as Function Approximaters}
We begin by formulating Transformers as function approximators to clarify their underlying rationale; this will also facilitate a better understanding of our parallelism solution.

Suppose we have a ground-truth function $f$ that defines the mapping from input $\mathbf{X}$ to output $\mathbf{Y}$:
\begin{equation}
\mathbf{Y}=f(\mathbf{X}).
\end{equation}
In general, any deep model can be represented as an approximator  $G(\mathbf{X};\mathbf{W})$, where $G$ is a deep neural network parameterized by $\mathbf{W}$. 
The objective is to learn the parameters $\mathbf{W}$ for the optimization problem:
\begin{equation}
\mathbf{W}^*=\argmin_{\mathbf{W}} \|G(\mathbf{X};\mathbf{W}) - f(\mathbf{X})\|.
\end{equation}

\begin{figure}[htbp!]
\centering
\includegraphics[width=0.48\textwidth]{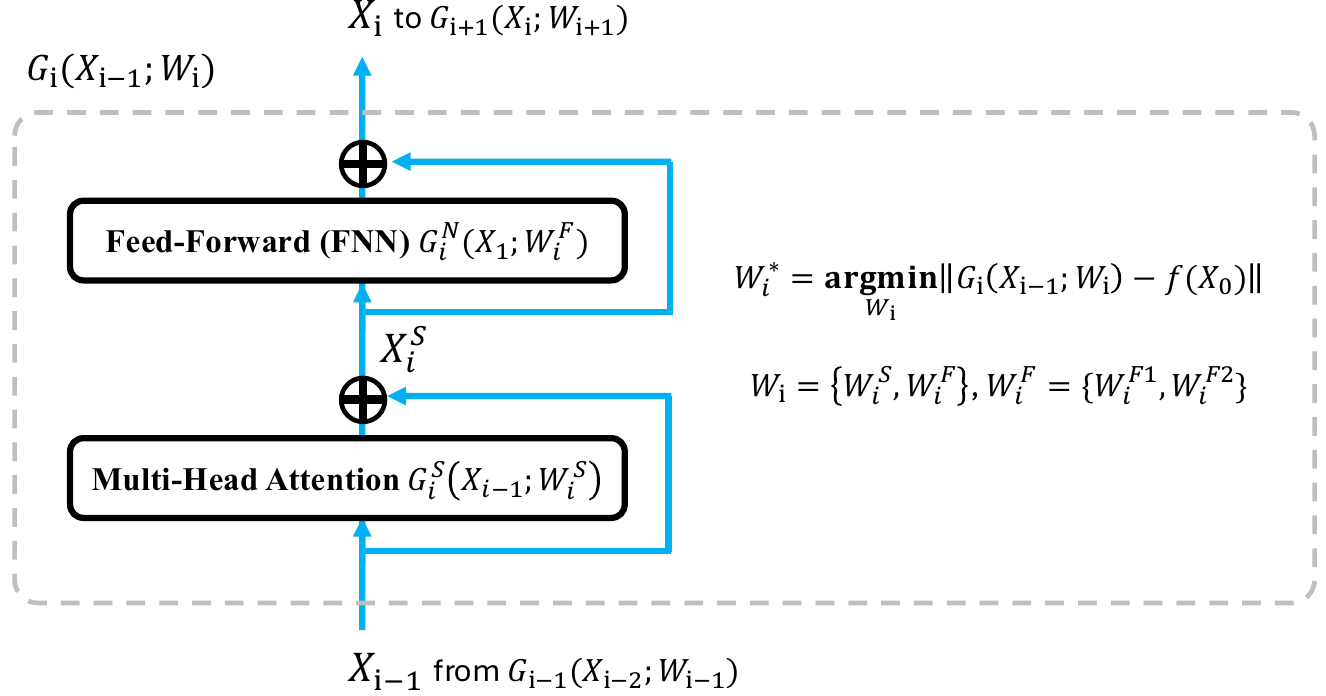}
\caption{The $i^{th}$ layer of a Transformer illustrated using the notations defined in this paper.}
\label{fig:trans_layer}
\end{figure}

\noindent\textbf{The Sequential Nature:} A Transformer implementes the function $G$ as a sequential architecture, stacking $n$ layers composed of self-attention and feed-forward neural network (FNN) blocks (Figure~\ref{fig:trans_layer}):
\begin{gather}
    G(\mathbf{X};\mathbf{W})=G_n(\cdots G_i(\cdots G_1(\mathbf{X}_1;\mathbf{W}_1)\cdots;\mathbf{W}_i) \cdots;\mathbf{W}_n))\nonumber\\
    \mathbf{X}_i=G_{i}(\mathbf{X}_{i-1};\mathbf{W}_{i})=G_i^{F}(G_i^{S}(\mathbf{X}_{i-1};\mathbf{W}_{i}^{S});\mathbf{W}_{i}^{F})\nonumber\\
    \mathbf{X}_0=\mathbf{X},\,\mathbf{W}=\{\mathbf{W}_i\}_1^n,\, n\in \mathbb{Z}^+
    \label{eq:transformer_whole}
\end{gather}
%
where the subscripts $1, i, n$ denote the layer indices, and the superscripts $S,N$ are block indicators. 
The functions $G^S$ and $G^F$ correspond to the approximations implemented by the self-attention and FFN blocks, respectively.

\noindent\textbf{Multi-Head Attention Block Formulation:} Within each layer, the single head attention block is implemented as
\begin{equation}
G_{i,j}^S(\mathbf{X}_{i-1})=softmax\left(\frac{\mathbf{Q}_{i,j}\mathbf{K}_{i,j}}{\sqrt{d}}\right)\mathbf{V}_{i,j}
\end{equation}
where $\mathbf{Q}_{i,j}, \mathbf{K}_{i,j}, \mathbf{V}_{i,j}$ are query, key, and value embeddings of the $j$-th head, respectively.
Letting $\mathbf{H}_{i,j}=softmax\left(\frac{\mathbf{Q}_{i,j}\mathbf{K}_{i,j}}{\sqrt{d}}\right)$ and $\mathbf{V}_{i,j}=\mathbf{X}_{i-1}\mathbf{W}_{i,j}^V$ ($\mathbf{W}_{i,j}^V$ is the is the value projection matrix), we have $G_{i,j}^S(\mathbf{X}_{i-1})=\mathbf{H}_{i,j}\mathbf{X}_{i-1}\mathbf{W}_{i,j}^V$.
By cancatating all heads, the whole multi-head process can then be written as 
\begin{equation}
    G_{i}^S(\mathbf{X}_{i-1}; \mathbf{W}_i)=\left[\underset{j=1} {\overset{h} {\LARGE ||}}\Big(\mathbf{H}_{i,j}\mathbf{X}_{i-1}\mathbf{W}_{i,j}^V\Big)\right]\mathbf{W}_i^O+\mathbf{X}_{i-1}
    \label{eq:self_atten_matrix}
\end{equation}
where ${||}$ denotes the cancatation operator,  $h$ is the total number of heads, and $\mathbf{W}_i^O$ is the weight matrix for fusing all heads.
For simplicity, we flatten all matrices $\mathbf{X}$ into their corresponding vector representations $\mathbf{x}$ and use these forms interchangeably from this point forward.
With this notation, Eq.~(\ref{eq:self_atten_matrix}) can be further transformed into 
\begin{gather}
\mathbf{x}_i^S=G_i^S(\mathbf{x}_{i-1};\mathbf{W}_i^S)=\mathbf{W}_i^S\mathbf{x}_{i-1}+\mathbf{x}_{i-1},\\
\mathbf{W}_i^S=\Big((\mathbf{W}_{i}^O)^\top \otimes \mathbf{I}\Big)\left[\underset{j=1} {\overset{h} {\LARGE ||}}(\mathbf{W}_{i,j}^V\otimes\mathbf{H}_{i,j}^\top)\right]^\top
\label{eq:self_attn_impl}
\end{gather}
where $\mathbf{x}_i^S$ is the output of the self-attention block, $\mathbf{I}$ is an unit matrix, and $\otimes$ is Kronecker multiplication operator.
%
%
Since the derivation of Eq.(\ref{eq:self_attn_impl}) involves a lengthy justification process, we present the conclusion directly here and omit the details for brevity.
The full derivation of this transformation is provided in Appendix.



\begin{figure*}[htbp!]
\centering
\includegraphics[width=0.98\textwidth]{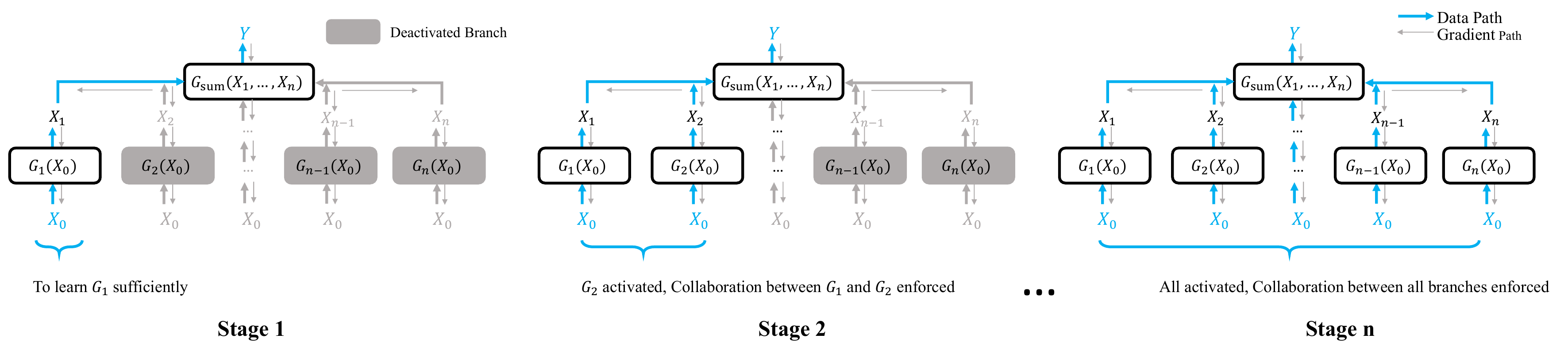}
\caption{The progressive training process in TeamFormer: new branches are incrementally activated to promote broader inter-branch collaboration once previous branches have been sufficiently trained.}
\label{fig:prog_training}
\end{figure*}

\noindent\textbf{FNN Block Formulation:} As the FNN block is inhertly a FCN network with the residual connection, it can written as
\begin{equation} \mathbf{X}_i=G_i^F(\mathbf{X}_i^S;\mathbf{W}_i^F)=\sigma(\mathbf{X}_i^S\mathbf{W}_i^{F1}+\mathbf{b}_i^{F1})\mathbf{W}_i^{F2} +\mathbf{b}_i^{F2}+\mathbf{X}_i^S,
\end{equation}
where $\mathbf{W}_i^F=\{\mathbf{W}_i^{F2},\mathbf{W}_i^{F1}, \mathbf{b}_i^{F1}, \mathbf{b}_i^{F2}\}$ is the paramter implementation, and $\mathbf{b}_i^{F1}$ and $\mathbf{b}_i^{F2}$ is the bais for the FNN block.
By substituting Eq.~(\ref{eq:self_attn_impl}), the calculation at layer $i$ can be reformulated as
\begin{equation} 
\begin{aligned}
\mathbf{x}_i=G_i(\mathbf{x}_{i-1};\mathbf{W}_i)&=\mathbf{W}_i^{F2}\sigma(\mathbf{W}_i^{1}\mathbf{x}_{i-1}  + \mathbf{b}_i^{F1}) +\mathbf{b}_i^{F2} \\
&+\mathbf{W}_{i}^S\mathbf{x}_{i-1} + \mathbf{x}_{i-1},
\end{aligned}
\end{equation}
where $\mathbf{W}_i^{1}=\mathbf{W}_i^{F1}\mathbf{W}_i^{S}+\mathbf{W}_i^{F1}$.
Let us denote $\hat{G}_i(\mathbf{x}_{i-1};\mathbf{W}_i)=\mathbf{W}_i^{F2}\sigma(\mathbf{W}_i^{1}\mathbf{x}_{i-1}  + \mathbf{b}_i^{F1}) +\mathbf{b}_i^{F2} +\mathbf{W}_{i}^S\mathbf{x}_{i-1}$.
As the learning goal at this layer $i$ is to minmize the gap between its output $G_i(\mathbf{x}_{i-1})$ and that of the target function $f(\mathbf{x})$, we have
\begin{align}
    \mathbf{W}_i^*=&\argmin_{\mathbf{W}_i} \|G_i(\mathbf{x}_{i-1};\mathbf{W}_i)-f(\mathbf{x}_0)\|\nonumber\\
    =&\argmin_{\mathbf{W}_i} \|\hat{G}_i(\mathbf{x}_{i-1};\mathbf{W}_i)  +\mathbf{x}_{i-1}-f(\mathbf{x}_0)\|\nonumber\\
    =&\argmin_{\mathbf{W}_i} \|\hat{G}_i(\mathbf{x}_{i-1};\mathbf{W}_i)-(f(\mathbf{x}_0)- \mathbf{x}_{i-1})\|\nonumber\\
    =&\argmin_{\mathbf{W}_i} \|\hat{G}_i(\mathbf{x}_{i-1};\mathbf{W}_i)-\hat{f}_i(\mathbf{x}_0,\mathbf{x}_{i-1})\|
\end{align}
where $\hat{f}_i(\mathbf{x}_0,\mathbf{x}_{i-1})=f(\mathbf{x}_0)- \mathbf{x}_{i-1}$.
Since $\mathbf{x}_{i-1}$ is the output of layer $i-1$ and is already computed and fixed by the time layer $i$ is reached, and $f(\mathbf{x}_0)$ is a fixed value (as $\mathbf{x}_0$ is given), the value of $\hat{f}_i(\mathbf{x}_0, \mathbf{x}_{i-1})$ is fixed. 
Thus, it can serve as a transformed target function for layer $i$.

Given that $\hat{G}_i(\mathbf{x}_{i-1}; \mathbf{W}_i)$ is a relaxed version of a fully connected network (proof provided in Appendix), the UAT introduced by \citet{Cybenko1989ApproximationBS} ensures the existence of $\hat{G}_i()$. 
This, in turn, guarantees the existence of ${G}_i()$.

By extending this reasoning to all $n$ layers, we derive a learning objective for the entire Transformer model as:
\begin{align}
    \mathbf{W}^*=&\argmin_{\mathbf{W}}\big\|f(\mathbf{x}_0)-\mathbf{x}_0-\sum_{i=1}^{n}\hat{G}^T_{i}(\mathbf{x}_{i-1})\big\|.
    \label{eq:multilayer_learning_goal}
\end{align}

From this formulation, we can identify two key properties of Transformers as
\begin{itemize}
    \item \textbf{Input-Output Difference Approximation}: Rather than viewing Transformers solely as output predictors, they can be interpreted as models that approximate the difference between the input and the output.
    
    \item \textbf{Collaborative Multi-Layer Approximation:} Each layer serves as an approximator for this difference, and the layers work together to progressively minimize it.
\end{itemize}
%

\section{Parallelized Shallow Transforms}

\noindent\textbf{Parallelized Difference Approximation}
Our parallelism solution, TeamFormer, offers a straightforward approach to implementing parallelized Transformers by leveraging the two identified properties.
Specifically, we can reduce inter-layer dependencies by training each layer independently as an individual approximator, $G_i(\mathbf{x}_0)$. 
The rationale is that, in a standard serial architecture, the transformed target at layer $i$, $\hat{f}_i(\mathbf{x}_0,\mathbf{x}_{i-1})=f(\mathbf{x}_0)- \mathbf{x}_{i-1}$, is a contineous function of $\mathbf{x}_0$.
We can thus use $\mathbf{x}_0$ in place of $\mathbf{x}_{i-1}$ and learn a $\hat{G}_{i}(\mathbf{x}_{0})$, which, by the UAT, can approximate this function as well. 
A detailed justification is provided in the Appendix due to space limitations.
This reformulates the learning objective in Eq.~(\ref{eq:multilayer_learning_goal}) into:
\begin{align}
\mathbf{W}^*=\argmin_{\mathbf{W}}\|f(\mathbf{x}_0)-\mathbf{x}_0-\sum_{i=1}^{n}\hat{G}_{i}(\mathbf{x}_{0})\|.
\label{eq:relax_learning_goal}
\end{align}
This relaxation enables parallel training of each layer and allows for arranging them in a shallow architecture, as illustrated in Figure~\ref{fig:prog_training}.
However, this approach disrupts the collaborative approximation property by removing the interactions between layers.

\noindent\textbf{Learning in Sequential Transformers} 
Let us first examine how multi-layer collaboration is achieved in standard Transformers. 
In the original sequential architecture, this collaborative property is inherently preserved, as each layer incrementally refines the approximation based on the output from preceding layers.
However, in practice, all layer parameters are optimized jointly with respect to the overall loss $\mathcal{L}(\cdot)=\mathcal{L}(G(\mathbf{x}_0);\mathbf{W}),f(\mathbf{x}_0))$, as given by
\begin{equation}
    \mathbf{W}^*=\argmin_{\mathbf{W}=\{\mathbf{W}_1,\mathbf{W}_2, \cdots, \mathbf{W}_n\}} \mathcal{L}(\cdot).
    \label{eq:simu_optm}
\end{equation}

\begin{algorithm}[H]
\caption{Progressive Approximation Training by Incrementally Enforcing Inter-branch Collaboration}
\label{alg:TeamFormer}
\begin{algorithmic}
\State TeamFormer with $n$ branches: $\{G_i,\mathbf{W}_i\}_1^n$;\\ Linear Aggregator: $G_{sum}$ with $\mathbf{W}_{sum}$;\\ Training data: $\{(\mathbf{X}, \mathbf{Y})\}$; Loss function: $\mathcal{L}$. 
\State
\For{epoch $= 1$ \textbf{to} $N_{\text{epochs}}$}
        \For{each batch $(\mathbf{X}_b, \mathbf{Y}_b)\subset\{(\mathbf{X}, \mathbf{Y})\}$}
            \For{$i = 1$ \textbf{to} $n$},
                \State $\mathbf{X}_0 =\textbf{Embedding}(\mathbf{X}_b)$
                \For{$j = 1$ \textbf{to} $i$} \Comment{Collaboration at stage $i$}
                    \State $\mathbf{X}_j \leftarrow G_j(\mathbf{X}_{0};\mathbf{W}_j)$ 
                \EndFor
                \State $\hat{\mathbf{Y}}_i \leftarrow G_{sum}([\mathbf{X}_1\cdots \mathbf{X}_i];\mathbf{W}_{sum})$
                \State $\mathcal{L}_i \leftarrow \mathcal{L}(\hat{\mathbf{Y}}_i, \mathbf{Y}_b)$
                \State Backpropagate $\mathcal{L}_i$ and update $\{\mathbf{W}_j\}_1^i$, $\mathbf{W}_{sum}$
            \EndFor
        \EndFor
    \For{each batch $(\mathbf{X}_b, \mathbf{Y}_b)$}
        \For{$i = 1$ \textbf{to} $n$}
            \State $\mathbf{X}_i \leftarrow G_i(\mathbf{X}_{b};\mathbf{W}_{i})$ 
        \EndFor
        \State $\hat{\mathbf{Y}} \leftarrow G_{sum}([\mathbf{X}_1\cdots \mathbf{X}_n];\mathbf{W}_{sum})$ \Comment{Prediction}
        \State $\mathcal{L} \leftarrow \mathcal{L}(\hat{\mathbf{Y}}, \mathbf{Y}_b)$
        \State Backpropagate $\mathcal{L}$ and update $\{\mathbf{W}_j\}_1^n$, $\mathbf{W}_{sum}$
    \EndFor
\EndFor
\end{algorithmic}
\end{algorithm}

\begin{table*}[h]
\centering
\caption{Performance comparison between TeamFormer (PF) variants and ViT$^{12}$ and ViT$^{24}$ on CIFAR-10, CIFAR-100, and Fashion-MNIST. The best performances are highlighted in bold, while results that surpass ViT$^{12}$ and ViT$^{24}$ are underlined.}
\resizebox{\textwidth}{!}{
\begin{tabular}{llc|c|c|c|c|c|c|c|c|c|c|c} 
\toprule
\multirow{3}{*}{\textbf{Benchmarks}}         &     & \multicolumn{6}{c|}{\textbf{Total Number of Layers = 12}}                       & \multicolumn{6}{c}{\textbf{Total Number of Layers = 24}}                         \\ 
\cmidrule(r){3-8}\cmidrule(l){9-14}
                              & \textbf{Model}    & \cellcolor{gray!10}\textbf{ViT}$^{12}$    & \textbf{TF}$_{12}^1$    & \textbf{TF}$^2_{6}$     & \textbf{TF}$^3_{4}$     & \textbf{TF}$^4_{3}$     & \textbf{TF}$^6_{2}$     & \cellcolor{gray!10}\textbf{ViT}$^{24}$    & \textbf{TF}$^1_{24}$     & \textbf{TF}$^2_{12}$    & \textbf{TF}$^4_{6}$     & \textbf{TF}$^6_{4}$     & \textbf{TF}$^{12}_{2}$     \\ 
\midrule
\multirow{2}{*}{\textbf{CIFAR-10}}      & \textbf{Acc.}($\uparrow$)      & \cellcolor{gray!10}88.48 & 78.40 & 85.86 & \underline{88.61} & 88.41 & \underline{88.76} & \cellcolor{gray!10}86.94 & 79.91 & 86.77 & \underline{88.76} & \textbf{89.85} & \underline{89.38}  \\
                              & \textbf{Loss}($\downarrow$)     & \cellcolor{gray!10}0.466 & 0.668 & 0.506 & 0.402 & 0.413 & 0.419 & \cellcolor{gray!10}0.617 & 0.654 & 0.542 & 0.478 & 0.423 & 0.450  \\ 
\midrule
\multirow{2}{*}{\textbf{CIFAR-100}}     & \textbf{Acc.}($\uparrow$)     & \cellcolor{gray!10}62.62 & 52.65 & 62.06 & \underline{64.19} & \underline{63.99} & \underline{64.54} & \cellcolor{gray!10}60.80 & 53.83 & \underline{62.62} & \underline{65.33} & \textbf{65.79} & \underline{62.53}  \\
                              & \textbf{Loss}($\downarrow$)     & \cellcolor{gray!10}1.826 & 2.191 & 1.755 & 1.632 & 1.617 & 1.609 & \cellcolor{gray!10}2.211 & 2.484 & 2.089 & 1.847 & 1.798 & 1.998  \\ 
\midrule
\multirow{2}{*}{\textbf{Fashion-MNIST}} & \textbf{Acc.}($\uparrow$)      & \cellcolor{gray!10}93.90 & 91.75 & 93.40 & \underline{93.94} & \underline{94.08} & \underline{93.95} & \cellcolor{gray!10}93.46 & 91.84 & \underline{93.55} & \textbf{94.48} & \underline{94.36} & \underline{94.22}  \\
                              & \textbf{Loss}($\downarrow$)     & \cellcolor{gray!10}0.229 & 0.229 & 0.189 & 0.182 & 0.186 & 0.197 & \cellcolor{gray!10}0.289 & 0.231 & 0.193 & 0.205 & 0.219 & 0.243  \\
\bottomrule
\end{tabular}}
\label{tab:Fea_Para_Former}
\end{table*}

If we assess the loss at an individual layer $i$ by computing $\mathcal{L}_i(G_i(\mathbf{x}_{i-1}), f(\mathbf{x}_0)) = \|f(\mathbf{x}_0) - \mathbf{x}_0 - G_i(\mathbf{x}_{i-1})\|$, and define learning at a layer as \textbf{progressive} if and only if $\mathcal{L}_i-\mathcal{L}_{i-1} > 0$, it becomes apparent that joint optimization across all layers does not necessarily guarantee progressive learning at each layer, since only the final composite loss is minimized.
In other words, while the sequential structure encourages inter-layer collaboration, it does not ensure that every layer consistently contributes in a productive direction. 
In fact, degradation can occur in certain layers when $\mathcal{L}_i-\mathcal{L}_{i-1} \leq 0$.

\noindent\textbf{Progressive Learning for Collaborative Approximation}
To preserve the collaborative effect in TeamFormer, we introduce a progressive learning paradigm.
Initially, all branches (corresponding to layers in the standard sequential Transformer) are deactivated. 
Training begins with only one active branch; once it is sufficiently trained, an additional branch is activated for further training. 
This process repeats, activating one new branch at a time.
It can be formulated as
\begin{equation}
\mathbf{W}^*=\argmin_{\mathbf{W}_1,\cdots,\mathbf{W}_n}\bigg(\cdots\Big(\argmin_{\mathbf{W}_1,\mathbf{W}_2}\big(\argmin_{\mathbf{W}_1} \mathcal{L}(\cdot)\big)\Big)\cdots\bigg).
    \label{eq:progr_optm}
\end{equation}

%
The training process is presented in Algorithm~\ref{alg:TeamFormer}.

This progressive training strategy effectively emulates the inter-layer collaboration inherent in the sequential model.
Moreover, it extends the concept of progressive learning beyond the capabilities of the original architecture.
Specifically, the simultaneous optimization process has been reformulated into a multi-stage optimization approach. 
In this approach, the optimization of $G_i()$ (with the further inclusion of the parameters $\mathbf{W}_i$) at each stage is performed only after ensuring that all preceding branches (i.e., $G_1()$ to $G_{i-1}()$) have been progressively trained, meaning the parameters $\mathbf{W}_1$ to $\mathbf{W}_{i-1}$ have already been optimized.
This approach adheres more closely to the theoretical analysis presented in the previous section, which establishes that the existence of $G_i()$ relies on the successful optimization of its predecessors as
\begin{equation}
    \left(\left(\left((\exists G_1\land\mathbf{W}^*_1\Rightarrow \exists G_2) \land \mathbf{W}^*_2\right) \cdots \right)\Rightarrow \exists G_n\right)\land\mathbf{W}^*_n\Rightarrow G()\nonumber
\end{equation}
Such a condition is not enforced in the simultaneous optimization of the original Transformers (i.e., Eq.~(\ref{eq:simu_optm})).

\section{Experiment}

\subsection{Experimental Settings}
We conduct extensive experiments by comparing our approach with ViT \cite{Dosovitskiy2020AnII}, one of the most representative Transformer models.
Both the 12-layer and 24-layer versions of ViT are used in the experiments, denoted as ViT$^{12}$ and ViT$^{24}$, respectively.
The multi-head attention block in ViT$^{12}$ and ViT$^{24}$ employs 3 attention heads, each with a dimension of 192. 
The feed-forward network (FFN) block uses a hidden layer with a dimension of 768. 
The experiments are conducted on three widely adopted benchmarks: CIFAR-10, CIFAR-100 \cite{Krizhevsky2009LearningML}, and Fashion-MNIST \cite{Xiao2017FashionMNISTAN}. We follow the standard protocol for splitting the training and testing subsets as defined by the respective benchmarks.
All models are trained for 300 epochs.
\textbf{All source code will be released on GitHub upon acceptance.}

\subsection{Can the parallel TeamFormer achieve performance comparable to standard sequential Transformers?}
To explore this, we vary the number of branches and the number of layers (depth) in each branch, ranging from 1 to 24. 
This results in 10 TeamFormer variants, designed to match the parameter scales of either ViT$^{12}$ or ViT$^{24}$.
For clarity, we use subscripts and superscripts to indicate the number of branches and the number of layers per branch, respectively, for these variants.
For example, TeamFormer$^4_6$ (or PF$^4_6$) represents a TeamFormer with 4 layers per branch and 6 branches.

The results are presented in Table~\ref{tab:Fea_Para_Former}.
Surprisingly, more than half of the TeamFormer variants outperform their counterparts, ViT$^{12}$ and ViT$^{24}$.
This superiority is consistently observed across all three benchmarks, with performance gains over ViT$^{12}$ (ViT$^{24}$) by $0.205\%\pm 0.0056$ ($2.390\%\pm 0.4463$), $1.586\%\pm 0.2656$ ($3.260\%\pm 1.4931$), and $0.080\%\pm 0.0725$ ($0.692\%\pm 0.3598$) on CIFAR-10, CIFAR-100, and Fashion-MNIST, respectively.
TeamFormer$^6_4$ achieves the best performance on 2 out of 3 benchmarks, outperforming all other counterparts and variants.
Meanwhile, TeamFormer$^3_4$ strikes a balance between performance (outperforming both ViT$^{12}$ and ViT$^{24}$, as well as most other variants, including those with 24 layers) and parallelism (with 4 branches).
These results validate the effectiveness of the parallelized architecture.

\begin{figure}[htbp!]
\centering
\includegraphics[width=0.5\textwidth]
{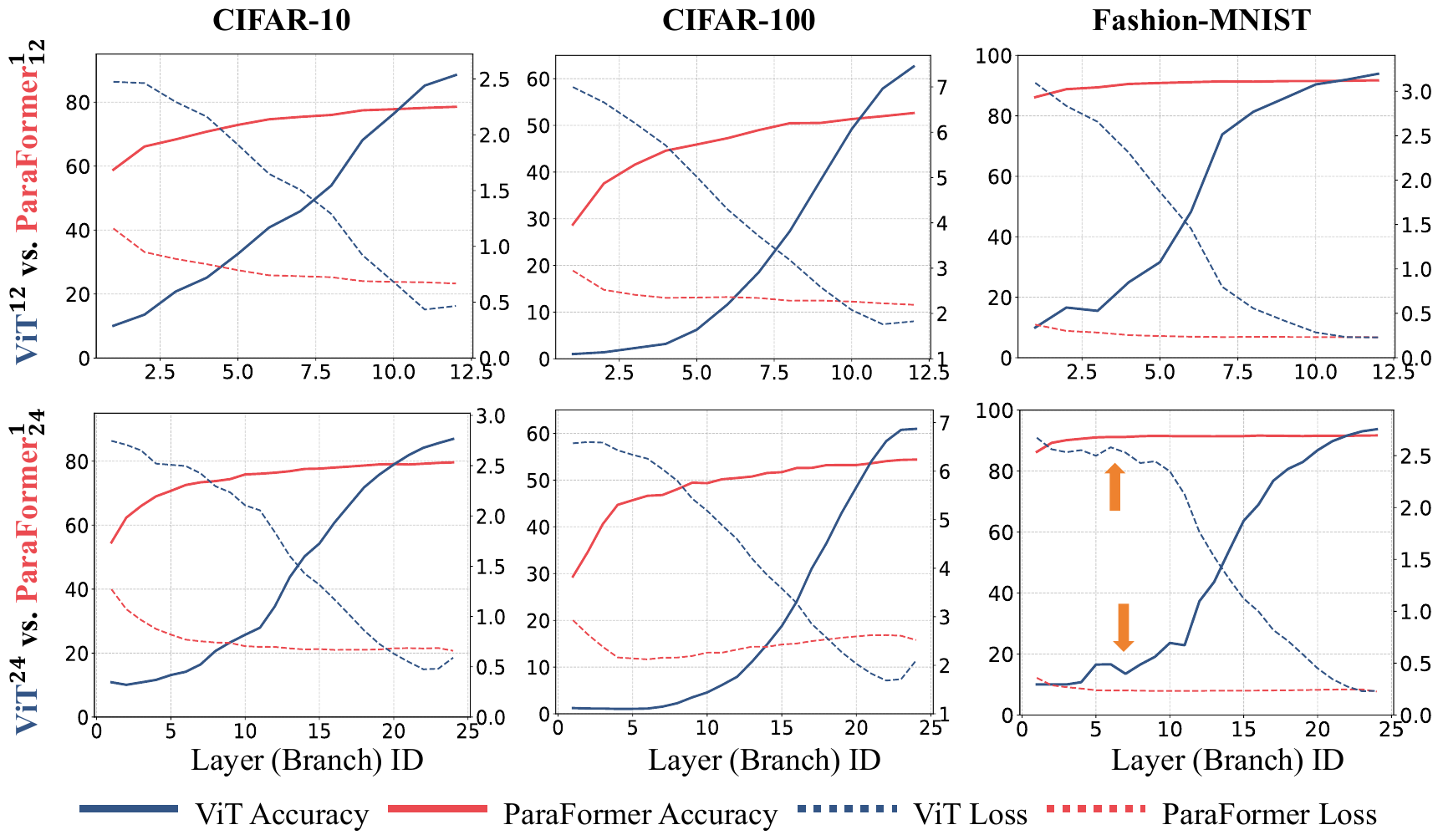}
\caption{The layer(branch)-wise accuracy and loss curves of ViT and TeamFormer on CIFAR-10, CIFAR-100 and Fashion-MNIST.}
\label{fig:layer_acc_loss_curvs}
\end{figure}

\begin{figure}[htbp!]
\centering
\includegraphics[width=0.48\textwidth]{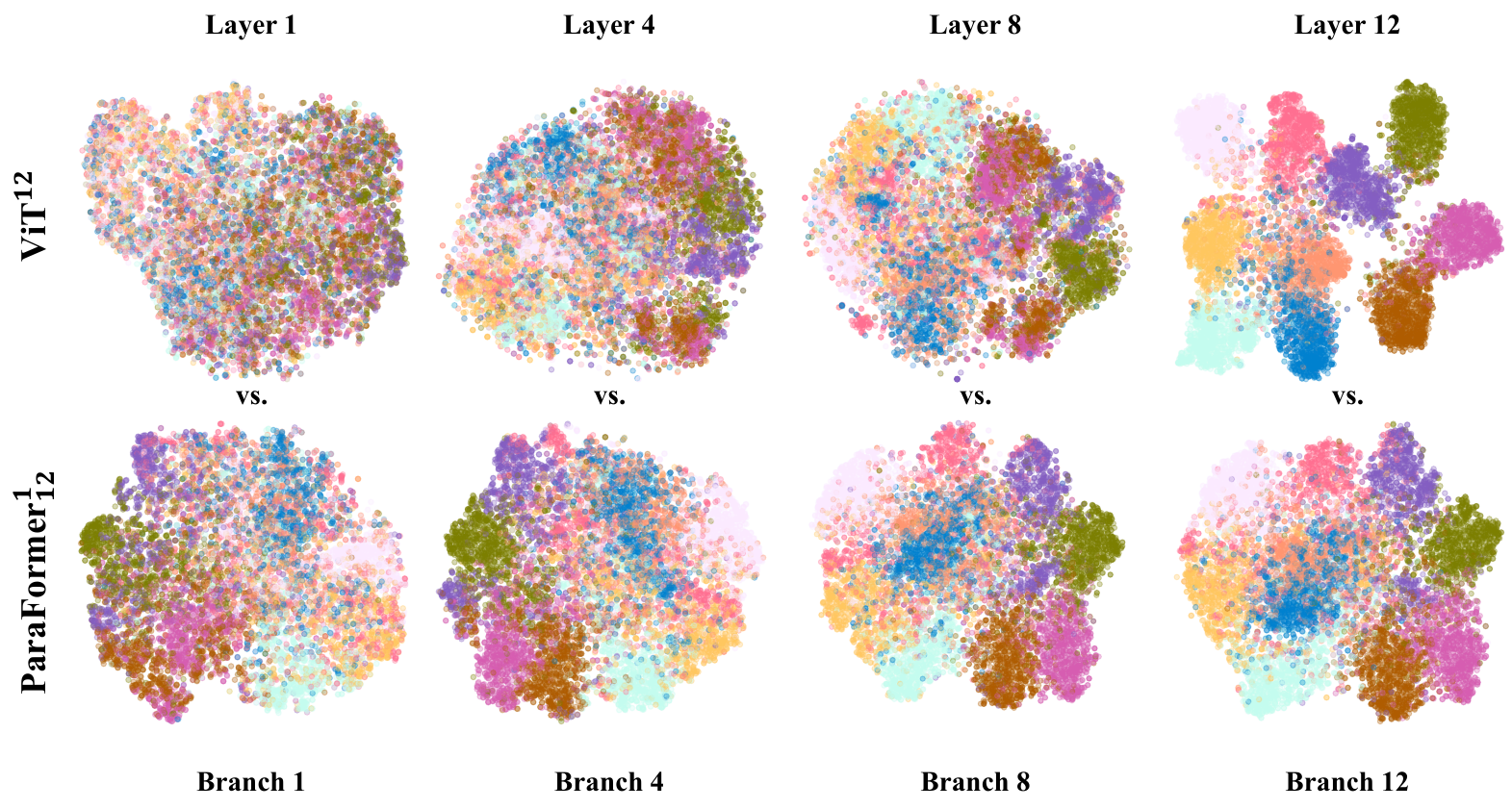}
\caption{Layer(Branch)-wise feature distributions of ViT$^{12}$ and TeamFormer$^1_{12}$ on CIFAR-10. TeamFormer$^1_{12}$ achieves satisfactory representations earlier than ViT$^{12}$.}
\label{fig:tsne_12_cifar10}
\end{figure}

\begin{figure}[htbp!]
\centering
\includegraphics[width=0.48\textwidth]{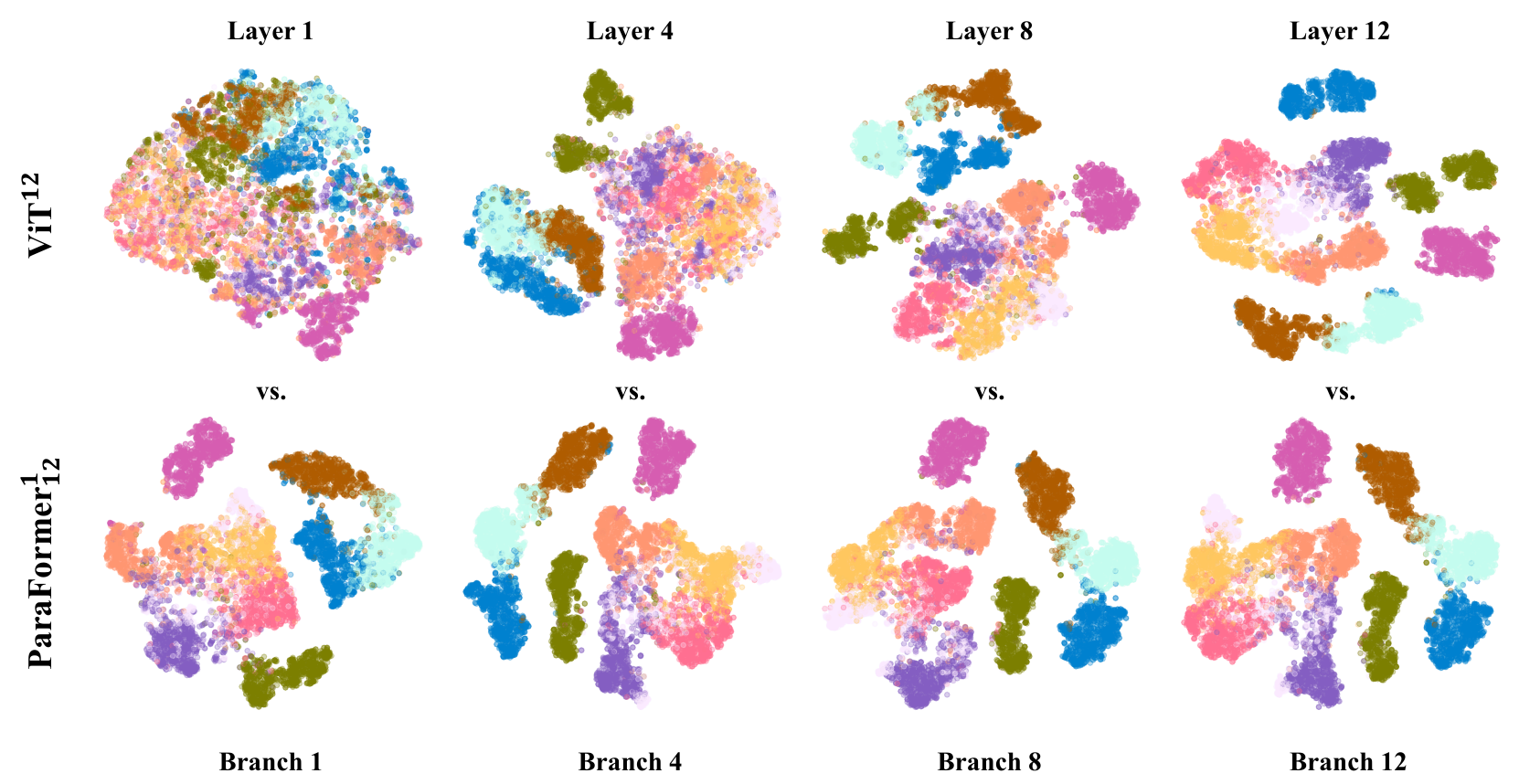}
\caption{Layer(Branch)-wise feature distributions of ViT$^{12}$ and TeamFormer$^1_{12}$ on Fashion-MNIST. TeamFormer$^1_{12}$ achieves satisfactory representations earlier than ViT$^{12}$.}
\label{fig:tsne_12_mnist}
\end{figure}

\subsection{Have inter-layer collaboration and progressive approximation been enforced?}
Our theoretical analysis highlights that successful learning relies on enforcing inter-layer collaboration and progressive approximation.
To evaluate how these principles are implemented in deep architectures (ViT$^{12}$ and ViT$^{24}$) versus shallow architectures (TeamFormer$^1_{12}$ and TeamFormer$^1_{24}$), we compute the accuracy and loss at each intermediate layer or branch and visualize the results in Figure~\ref{fig:layer_acc_loss_curvs}.

\noindent\textbf{Inter-Layer Collaboration:} The results indicate that both deep and shallow architectures enforce collaboration, as evidenced by the consistent increase in accuracy (and decrease in loss) with the addition of more layers or branches.
This demonstrates that depth is not the only means of achieving collaboration, validating the effectiveness of our proposed algorithmic-level enforcement in TeamFormer.

\noindent\textbf{Progressive Approximation:} TeamFormer demonstrates a more effective implementation of progressive approximation, as shown by its competitive accuracy when only a single branch is used (e.g., $86.17\%$ and $86.30\%$ at branch 1 on Fashion-MNIST). 
Additionally, its accuracy steadily improves toward the optimal level as more branches are incorporated.
In contrast, ViT, with its standard sequential architecture, does not exhibit this property, as progressive approximation is not explicitly enforced.
Instead, its inter-layer collaborative approximation is more flexible, which can lead to instances where later layers negate the contributions of earlier layers. 
For example, on Fashion-MNIST, its accuracy at layer 7 drops compared to that of layer 6.

This is also evident at the representation level.
In Figure~\ref{fig:tsne_12_cifar10} and Figure~\ref{fig:tsne_12_mnist}, we use the outputs from each layer/branch as feature maps to plot t-SNE distributions for the classes in CIFAR-10 and Fashion-MNIST.
It is clear that TeamFormer represents the classes more effectively at earlier stages compared to ViT.
In contrast, ViT features mix the classes until nearing the final layer.

\subsection{What advantages does progressive approximation offer for model compression and expansion?}

\noindent\textbf{Model Compression:} As shown in Figure~\ref{fig:layer_acc_loss_curvs}, the performance of TeamFormer converges well before all branches are utilized.
For instance, TeamFormer$^1_{24}$ achieves an accuracy of $91.03\%$ at branch 5, which is very comparable to the $91.67\%$ accuracy at the fully-loaded stage with 24 branches.
In many applications, this level of performance is sufficient enough, allowing us to discard the later branches and compress the model by up to $4.8\times$.
This significantly lowers the computational requirements, enabling deployment in scenarios where clients lack the resources to run a fully-loaded model.
Note that this is not to present a state-of-the-art compression method, but to provide a \textbf{starting point} that is orthogonal and compatible with existing techniques such as quantization, pruning, and distillation. 
For example, by applying quantization (from float32 to float8) to TeamFormers, it yields an additional $3.78\times$ compression, resulting in a total compression up to $15.07\times$.

\begin{figure}[htbp!]
\centering
\includegraphics[width=0.4\textwidth]{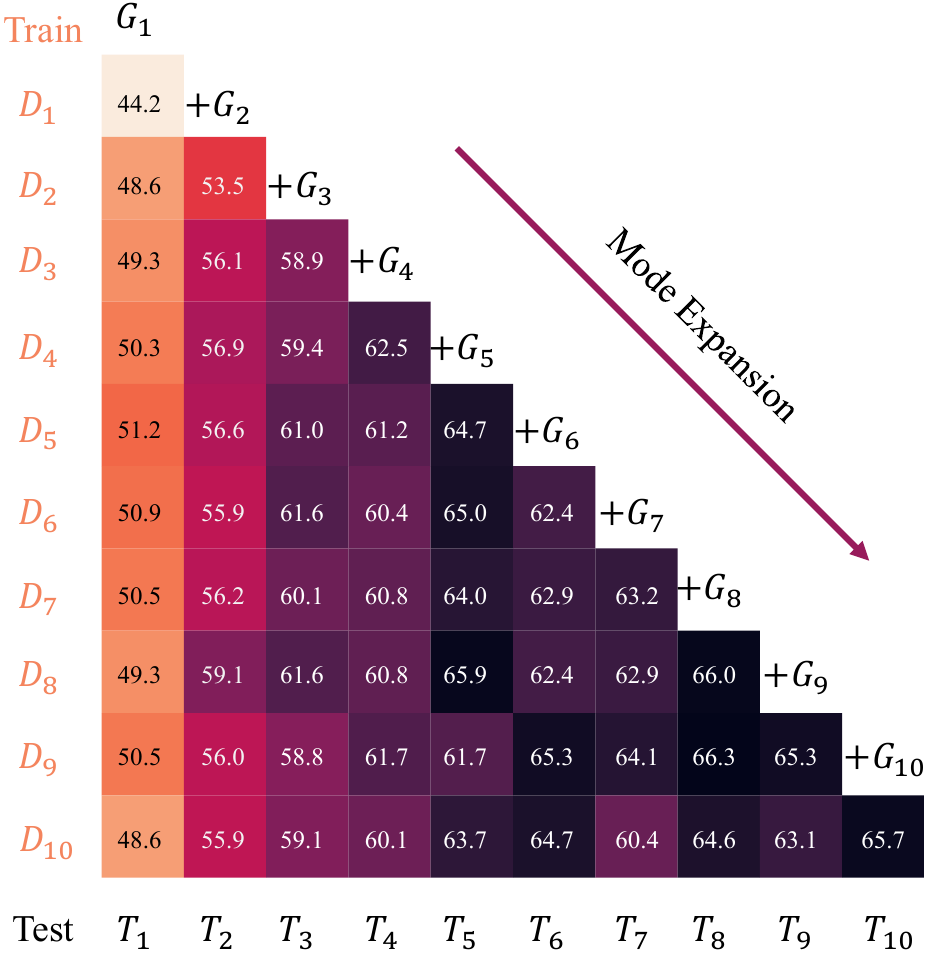}
\caption{Model expansion experiment on CIFAR-10: Training begins with a single branch $G_1$ (TeamFormer$_1^1$), and additional branches are incrementally added one at a time, until the model reaches 10 branches (TeamFormer$_{10}^1$).}
\label{fig:model_expan}
\end{figure}

\noindent\textbf{Modal Expansion:}  To simulate scenarios where a model needs to expand to incorporate new data, we divide the CIFAR-10 dataset into 10 equal subsets, denoted as ${(D_i, T_i)}_i^{10}$, where each $(D_i, T_i)$ represents a training and testing pair.
We begin by training and testing a TeamFormer model with a single branch $G_1$ on $(D_1, T_1)$.
Next, we introduce a new branch $G_2$ and train/test it using $(D_2, T_2)$.
This process is repeated, adding one new branch at each step.
To assess whether expanding the model and further training affects the performance of previous branches, we also evaluate each branch on its respective original test set (e.g., $G_1$ on $T_1$) to check for potential forgetting of previously learned knowledge.
The results are presented in Figure~\ref{fig:model_expan}.

Our findings reveal that TeamFormer possesses a highly extensible architecture, enabling updates with new data without requiring a complete model rebuild.
Furthermore, the performance of existing branches remains stable during expansion, as evidenced by the consistent colors across columns in the figure, indicating effective retention of previously learned knowledge.
Moreover, most performance blocks become darker during expansion, suggesting that the understanding of old knowledge in existing branches can be further improved through collaboration with newly added branches.
Again, this experiment is not meant to propose a state-of-the-art model expansion method, but to demonstrate that TeamFormer offers a foundation for approaches like continual learning.
TeamFormer is orthogonal to, and compatible with, such methods.

\begin{table}[h]
\centering
\caption{Inference efficiency comparison between Sequential Transformer, FairScale, GPiPe and TeamFormer ($X=12/N$) variants on CIFAR-10. Time are reported in milliseconds.}
\resizebox{0.48\textwidth}{!}{
\begin{tabular}{lcc|c|c|c} 
\toprule
& & \multicolumn{4}{c}{\textbf{Number of GPUs (N)}}\\
\cmidrule(){3-6}
\textbf{Method} & \textbf{Model} & 1 & 3 & 4 & 6\\ 
\midrule
{Sequential} &{ViT}$^{12}$ & \cellcolor{gray!10}{1572.9} & - & - & - \\
{FairScale}  &{ViT}$^{12}$ & - &\cellcolor{gray!10}{2768.5} & \cellcolor{gray!10}{3116.5} & \cellcolor{gray!10}{3749.3}  \\
{GPiPe} & {ViT}$^{12}$ & - &\cellcolor{gray!10}{2784.7}  & \cellcolor{gray!10}{3139.6} & \cellcolor{gray!10}{3804.8}  \\
\textbf{TeamFormer} & PM$^X_N$ & \textbf{1562.5} & \textbf{1159.8} & \textbf{975.8} &  \textbf{876.3}\\
\bottomrule
\end{tabular}}
\label{tab:Time_Comparison}
\end{table}

\subsection{How Efficient Is TeamFormer for Inference?}
To evaluate the efficiency of TeamFormer as a parallel computing solution, we compare its inference performance with FairScale (a popular parallelism framework developed by Facebook AI Research \cite{FairScale2021}) and GPiPe (a parallelism solution developed by Google \cite{Huang2018GPipeET}) .
The sequential solution using a single GPU has also been compared as a reference.
The experiment is conducted on an 8xA6000 workstation by deploying ViT$^{12}$ and TeamFormer variants across 1, 3, 4, and 6 GPUs, respectively.
Inference is performed on the CIFAR-10 dataset, and the total time to process all $10,000$ images is reported in milliseconds. 
The results are presented in Table~\ref{tab:Time_Comparison}.

The results indicate that TeamFormer is $330.33\%\pm 86$ faster than FairScale and GPipe, showcasing a significant advantage in inference efficiency.
In general, TeamFormer achieves higher speeds as more GPUs are utilized for large-scale parallelism.
Compared to the sequential approach on a single GPU, TeamFormer delivers an improvement of $284.47\pm71.92$ milliseconds by adding one GPU.

The efficiency of TeamFormer may offer significant advantages when applied to various scenarios. 
Its ability to leverage multiple GPUs while maintaining high inference speeds allows for faster model deployment and reduced latency, which is critical for time-sensitive applications like autonomous systems or online recommendation engines.
This efficiency may also translate to cost savings by maximizing GPU utilization and minimizing computation time, making TeamFormer an attractive choice for resource-intensive environments.

\section{Conclusion} 
In this paper, we introduced TeamFormer, a shallow Transformer architecture that achieves true parallelism in both structural and computational aspects. 
By formulating standard Transformers as function approximators, we demonstrated through theoretical analysis that their performance is rooted in inter-layer collaboration for progressive approximation, rather than depth itself. 
This finding highlights that while sequential designs in deep Transformers enforce such collaboration, it is not inherently tied to depth or sequentiality.  
TeamFormer removes the sequential constraints by organizing layers into parallel branches, algorithmically ensuring inter-layer collaboration. 
Moreover, it enforces progressive approximation, enabling faster convergence by ensuring each new branch further reduces the loss from preceding branches.  
Comprehensive experiments validate TeamFormer’s effectiveness, outperforming standard Transformers like ViT. 
Additionally, TeamFormer enables up to $480\%$ model compression, making it highly resource-efficient, and supports adaptive continuous learning for model expansion.  
These advancements are grounded in our theoretical formulation of Transformers based on the Universal Approximation Theorem, which not only demystifies the ``depth belief'' but also paves the way for designing more efficient and scalable Transformer architectures.

Though promising, TeamFormer is a newly introduced architecture that has received limited exploration to date.
As an initial pilot study, we implemented it using existing tools like PyTorch.
We believe that CUDA-level optimizations hold great potential to further enhance its performance, and we aim to explore this in future work.

\bibliography{main}
\bibliographystyle{icml2025}
\end{document}